\documentclass[10pt,twocolumn,letterpaper]{article}

\usepackage{wacv}
\usepackage{times}
\usepackage{epsfig}
\usepackage{amsmath}
\usepackage{amssymb}
\usepackage{mathtools}
\usepackage[ruled, vlined, linesnumbered]{algorithm2e}
\newlength\mylen

\usepackage{mathtools}
\usepackage{dsfont}
\usepackage{booktabs}
\usepackage{makecell}
\setcellgapes{3.5pt}
\usepackage{bm}
\usepackage[table,xcdraw]{xcolor}
\usepackage{graphicx}\graphicspath{{Figures/}{tiff/}}

\usepackage[font=small,justification=centering]{caption}

\wacvfinalcopy 

\def\wacvPaperID{238} 

\ifwacvfinal\fi
\begin{document}

\title{Video Jigsaw: Unsupervised Learning of Spatiotemporal Context for Video Action Recognition}

\author{Unaiza Ahsan \\
Georgia Institute of Technology\\
{\tt\small uahsan3@gatech.edu}
\and
Rishi Madhok \\
Carnegie Mellon University\\
{\tt\small rmadhok@andrew.cmu.edu}
\and
Irfan Essa \\
Georgia Institute of Technology\\
{\tt\small irfan@gatech.edu}
}

\maketitle
\ifwacvfinal\thispagestyle{empty}\fi

\begin{abstract}
   We propose a self-supervised learning method to jointly reason about spatial and temporal context for video recognition. Recent self-supervised approaches have used spatial context~\cite{doersch2015unsupervised,noroozi2016unsupervised} as well as temporal coherency~\cite{misra2016shuffle} but a combination of the two requires extensive preprocessing such as tracking objects through millions of video frames~\cite{wang2017transitive} or computing optical flow to determine frame regions with high motion~\cite{lee2017unsupervised}. We propose to combine spatial and temporal context in one self-supervised framework without any heavy preprocessing. We divide multiple video frames into grids of patches and train a network to solve jigsaw puzzles on these patches from multiple frames. So the network is trained to correctly identify the position of a patch within a video frame as well as the position of a patch over time. We also propose a novel permutation strategy that outperforms random permutations while significantly reducing computational and memory constraints. We use our trained network for transfer learning tasks such as video activity recognition and demonstrate the strength of our approach on two benchmark video action recognition datasets without using a single frame from these datasets for unsupervised pretraining of our proposed video jigsaw network.
\end{abstract}

\section{Introduction}

Unsupervised representation learning of visual data is a much needed line of research as it does not require manually labeled large scale datasets. Especially for classification tasks in video, where the annotation process is tedious and sometimes hard to agree upon (where an action begins and ends for example)~\cite{sigurdsson2017actions}. One proposed solution to this problem is self-supervised learning where auxiliary tasks are designed to exploit the inherent structure of unlabeled datasets and a network is trained to solve those tasks. Self-supervised tasks that exploit spatial context include predicting the location of one patch relative to another \cite{doersch2015unsupervised}, solving a jigsaw puzzle of image patches \cite{noroozi2016unsupervised}, predicting an image's color channels from grayscale \cite{zhang2016colorful,larsson2016learning} among others. Self-supervision tasks on video data include video frame tuple order verification \cite{misra2016shuffle}, sorting video frames \cite{lee2017unsupervised} and tracking objects over time and training a Siamese network for similarity based learning \cite{wang2015unsupervised}. Video data involves not just spatial context but also rich temporal structure in an image sequence. Attempts to combine the two have resulted in multi-task learning approaches \cite{doersch2017multi} that result in some improvement over a single network. This work proposes a self-supervised task that jointly exploits spatial and temporal context in videos by dividing multiple video frames into patches and shuffling them into a jigsaw puzzle problem. The network is trained to solve this puzzle that involves reasoning over space and time.

There are several studies that empirically validate that the earliest visual cues captured by infants' brains are surface motions of objects \cite{spelke1990principles}. These then go on and develop into perception involving local appearance and texture of objects. \cite{spelke1990principles}. Studies have also pointed out that objects' motion and their temporal transformation are important for the human visual system to learn the structure of objects \cite{foldiak1991learning,wiskott2002slow}. Motivated by these studies, there is recent work on unsupervised video representation learning via tracking objects through videos and training a Siamese network to learn a similarity metric on these object patches \cite{wang2015unsupervised}. However, the prerequisite of this approach is to track millions of objects through videos and extract the relevant patches. Keeping this in mind, we propose to learn such a structure of objects and their transformations over time by designing a self-supervised task which solves jigsaw puzzles comprising multiple video frame patches, without needing to explicitly track objects over time. Our proposed method, trained on a large scale video activity dataset also does not require optical flow based patch mining and we show empirically that a large unlabeled video dataset with a simple permutation sampling approach are enough to learn an effective unsupervised representation. Figure~\ref{fig1} shows our proposed approach, which we call \textit{video jigsaw}. Our contributions in this paper are:
\vspace{-2mm}

\begin{figure}[t]
\centering
\includegraphics[height=6.5cm]{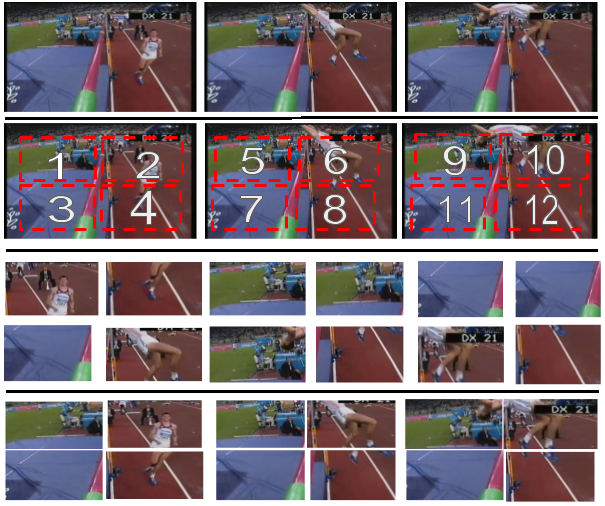}
\caption{Video Jigsaw Task: The first row shows a tuple of frames of action ``high jump''. Second row shows how we divide each frame into a 2x2 grid of patches. The third row shows a random permutation of the 12 patches which are input to the network. The final row shows the jigsaw puzzle assembled}
\label{fig1}
\end{figure}

\begin{enumerate}
 \item We propose a novel self-supervised task which divides multiple video frames into patches, creates jigsaw puzzles out of these patches and the network is trained to solve this task.
 \item Our work exploits both spatial and temporal context in one joint framework without requiring explicit object tracking in videos or optical flow based patch mining from video frames.
 \item We propose a permutation strategy that constrains the sampled permutations and outperforms random permutations while being memory efficient. 
 \item We show via extensive experimental evaluation the feasibility and effectiveness of our approach on video action recognition. 
 \item We demonstrate the domain transfer capability of our proposed video jigsaw networks, given that our best self-supervised model is trained on Kinetics \cite{kay2017kinetics} video frames and we demonstrate competitive results on UCF101 \cite{soomro2012ucf101} and HMDB51\cite{kuehne2011hmdb} datasets.
\end{enumerate}

\section{Related Work}
Unsupervised representation learning is a well studied problem in the literature for both images and videos. The goal is to learn a representation that is \textit{simpler} in some way: it can be low-dimensional, sparse, and/or independent \cite{goodfellow2016deep}. One way to learn such a representation is to use a reconstruction objective. Autoencoders \cite{hinton2006reducing} are neural networks designed to reconstruct the input and produce it as its output. Denoising autoencoders \cite{vincent2008extracting} train a network to undo random corruption of the input data. Other methods that use reconstruction to estimate the latent variables that can explain the observed data include Deep Boltzmann Machines \cite{salakhutdinov2010efficient}, stacked autoencoders \cite{lee2007efficient,bengio2007greedy} and Restricted Boltzmann Machines (RBMs) \cite{hinton1986learning,smolensky1986information}. Classical work (before deep learning) involved hand-designing features and feature aggregation for application such as object discovery in large datasets \cite{sivic2005discovering,russell2006using} and mid-level feature mining \cite{doersch2013mid,singh2012unsupervised,sun2013learning}. 

Unsupervised learning from videos include many learning variants such as video frame prediction \cite{wiskott2002slow,mobahi2009deep,srivastava2015unsupervised,taylor2010convolutional,goroshin2015unsupervised} but we argue that predicting pixels is a much harder task, especially if the end task is to learn high level motion and appearance changes in frames for activity recognition. Other unsupervised representation learning approaches include exemplar CNNs \cite{dosovitskiy2016discriminative}, CliqueCNNs \cite{bautista2016cliquecnn} and unsupervised similarity learning by clustering \cite{bautista2017deep}. 

Unsupervised representations are generally learned to make another learning task (of interest) easier \cite{goodfellow2016deep}. This forms the basis of another line of work that has emerged, called `self-supervised learning' \cite{doersch2015unsupervised,misra2016shuffle,zhang2016colorful,larsson2016learning,wang2015unsupervised,doersch2017multi,wang2017transitive,noroozi2017representation}. Self-supervised learning aims to find structure in the unlabeled data by designing auxiliary tasks and pseudo labels to learn features that can explain the factors of variation in the data. These features can then be useful for the target task; in our case, video action recognition. Self-supervised learning can exploit several cues, some of which are spatial context and temporal coherency. Other self-supervised learning tasks on videos use cues like ego-motion \cite{zhou2015temporal,jayaraman2015learning,agrawal2015learning} as a supervisory signal and other modalities beyond raw pixels such as audio \cite{owens2016ambient,owens2016visually} and robot motion \cite{agrawal2016learning,pinto2017supervision,pinto2016curious,pinto2017learning}. We briefly cover relevant literature from the spatial, temporal and combined contextual cues for self-supervised learning.
\vspace{-2mm}
\paragraph{\textbf{Spatial Context:}} These methods typically sample patches from images or videos. Supervised tasks are designed around the arrangement of these patches and pseudo labels constructed. Doersch \etal \cite{doersch2015unsupervised} divide an image into a 3x3 grid, sample two patches from an image and train a network to predict the location of the second patch relative to the first. This prediction task requires no labels but learns an effective image representation. Noroozi and Favaro \cite{noroozi2016unsupervised} also divide an image into a 3x3 grid but they input all patches in a Siamese-like network where the patches are shuffled and the task is to solve this jigsaw puzzle task. They report that with just 100 permutations, their network is able to learn a representation such that when finetuned on PASCAL VOC 2007 \cite{everingham2015pascal} for object detection and classification, it produces good results. Pathak \etal \cite{pathak2016context} devise an inpainting auxiliary task where blocks of pixels from an image are removed and the task is to predict the missing pixels. A related task is the image colorization one \cite{zhang2016colorful,larsson2016learning} where the network is trained to predict the color of the image which is available as a `free signal' with images. Zhang \etal \cite{zhang2017split} modify the autoencoder architecture to predict raw data channels as their self-supervised task and use the learnt features for supervised tasks. 
\vspace{-3mm}
\begin{figure*}[t]
\centering
\includegraphics[width=0.7\textwidth,height=6.5cm]{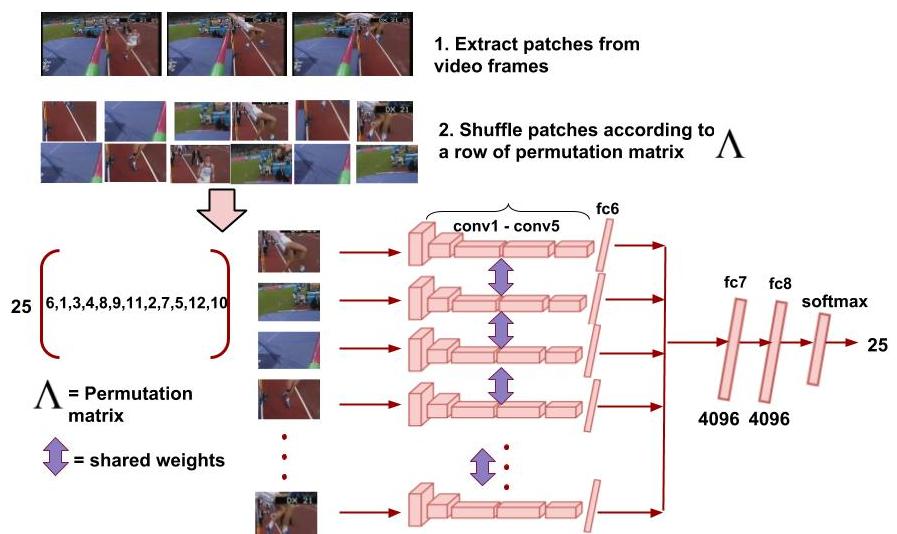}
\caption{Our full video jigsaw network training pipeline.}
\label{fig2}
\end{figure*}
\paragraph{\textbf{Temporal Coherency:}}
These methods use temporal coherency as a supervisory signal to train models and use abundant unlabeled video data instead of just images. Wang and Gupta \cite{wang2015unsupervised} use detection and tracking methods to extract object patches from videos and train a Siamese network with the prior that objects in nearby frames are similar whereas other random object patches are dissimilar. Misra \etal \cite{misra2016shuffle} devise a sequence verification task where tuples of video frames are shuffled and the network is trained on the binary task of discriminating between correctly ordered and shuffled frames. Fernando \etal \cite{fernando2017self} design a task where they take frames in correct temporal order and shuffled order, encode them and pass them as input to a network which is then trained to predict the odd encoding out of the rest; odd being the temporally shuffled one. Lee \etal \cite{lee2017unsupervised} extract high motion tuples of four frames via optical flow and shuffle them. Their network learns to predict the permutation from which the frames were sampled from. Our work is highly related to approaches that shuffle video frames and train a network to learn the permutations. A key difference between our work and Lee \etal \cite{lee2017unsupervised} is that they use only a single 80 x 80 patch from a video frame and shuffle it with three other patches from different frames. We sample a grid of patches from each frame and shuffle them with other multiple patches from other frames. Instead of the binary task of tuple verification like Misra \etal \cite{misra2016shuffle}, our self-supervised task is to predict the exact permutation of the patches, much like the jigsaw puzzle task of Noroozi and Favaro \cite{noroozi2016unsupervised} --- only on videos. Some recent approaches have used temporal coherency-based self-supervision on video sequences to model fine-grained human poses and activities \cite{milbich2017unsupervised} and animal behavior \cite{brattoli2017lstm}. Our model is not specialized for motor skill learning like \cite{brattoli2017lstm} and we do not require bounding boxes for humans in the video frames as in \cite{milbich2017unsupervised}. 
\vspace{-2mm} 
\paragraph{\textbf{Combining Multiple Cues:}}
Since our approach combines spatial and temporal context into a single task, it is pertinent to mention recent approaches to combine multiple supervisory cues. Doersch and Zisserman \cite{doersch2017multi} combine four self-supervised tasks in a multi-task training framework. The tasks include context prediction \cite{doersch2015unsupervised}, colorization \cite{zhang2016colorful}, exemplar-based learning \cite{dosovitskiy2014discriminative} and motion segmentation \cite{pathak2017learning}. Their experiments prove that naively combining different tasks does not yield improved results. They propose a lasso regularization scheme to capture only useful features from the trained network. Our work does not require a complex model for combining the spatial and temporal context prediction tasks for self-supervised learning. Wang \etal \cite{wang2017transitive} train a Siamese network to recognize if an object patch belongs to a similar category (but different object) or it belongs to the same object, only later in time. This work attempts to combine spatial and temporal context but requires preprocessing to discover the tracked object patches. Our work constructs the spatiotemporal task from video frames automatically without requiring graph construction or visual detection and tracking. There is also recent work on using synthetic imagery and its `free annotations' to learn visual representations \cite{ren2017cross} by combining multiple self-supervised tasks. A related approach to ours is that of \cite{sumer2017self} where the authors devise two tasks for the network to train on in a multi-task framework. One is spatial placement task where a network learns to identify if a an image patch overlaps with a person bounding box or not. The second task is an ordering one where a network is trained to identify the correct sequence of two frames in a Siamese network setting much like \cite{misra2016shuffle}. The key difference between their work and ours is that our network does not do multi-task learning and predicts a much richer set of labels (that is, the shuffled configuration of patches) as compared to binary classification.


\section{The Video Jigsaw Puzzle Problem}\label{sec-train}
We present the video jigsaw puzzle task in this section. Our goal is to create a task that not only forces a network to learn part-based appearance of complex activities but also, how those parts change over time. For this, we divide a video frame into $2 \times 2$ grid of patches. For a tuple of three video frames, this results in $3 \times (2 \times 2) = 12$ total patches per video. We number the patches from $1$ to $12$ and shuffle them. Note that there are $12! = 479001600$ ways to shuffle these patches. We use a small but diverse subset of these patches' permutations, selecting them based on their Hamming Distance from the previously sampled permutations \cite{noroozi2016unsupervised}. We use two sampling strategies in our experiments which we will describe in more detail. The network is trained to predict the correct order of patches. Our video jigsaw task is illustrated in Figure~\ref{fig1}. 

\subsection{Training Video Jigsaw Network}
Our training strategy follows a line of recent works on self-supervised learning on large scale image and video datasets \cite{noroozi2016unsupervised,lee2017unsupervised}. Typically, the self-supervised task is constructed by defining pseudo labels --- in our case, the permuted order of patches. Then, each patch, after undergoing preprocessing, is input to a multi-stream Siamese-like network. Each stream, up till the first fully connected layer, shares parameters and operates independently on the frame patches. After the first fully connected layer (\textit{fc6}), the feature representations are concatenated and input to another fully connected layer (\textit{fc7}). The final fully connected layer transforms the features to a $N$ dimensional output, where $N$ is the number of permutations. A softmax over this output returns the most likely permutation the frame patches were sampled from. Our detailed training network is shown in Figure~\ref{fig2}.

\begin{figure*}[t]
\centering
\includegraphics[width=0.6\textwidth,height=7.5cm, keepaspectratio]{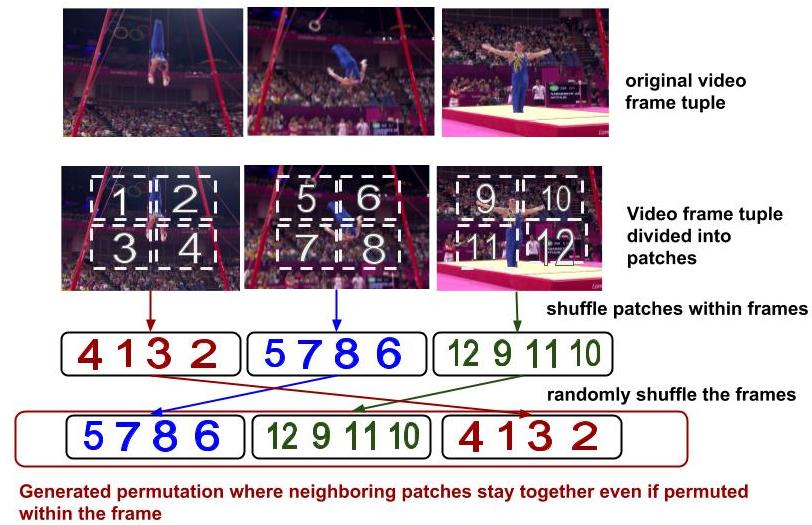}
\caption{Our proposed permutation sampling strategy. We randomly permute the patches within each frame in a tuple, then we permute the frames. Since the number of patches per frame is $4$, there are $4! = 24$ unique ways to shuffle these patches within a frame. We repeat this for all frames in the tuple and finally select the top $N$ permutations based on Hamming distance. This strategy preserves spatial coherence, preserves diversity between permutations, takes a fraction of the time and memory as compared to the algorithm of \cite{noroozi2016unsupervised} and results in either comparable or better performance in the transfer learning tasks}
\label{fig3}
\end{figure*}

\begin{algorithm}[ht]
\DontPrintSemicolon
\KwIn{Number of permutations $N$, patches per frame $n_p$, number of frames $n_f$}
\KwOut{Permutation Matrix $\Lambda$}
\SetKwBlock{Begin}{function}{end function}
\Begin($\textit{generatePerm} {(} N,n_p,n_f {)}$)
{
  \For{$i = 1:n_f $} 
  {
    $\lambda_{i} \leftarrow$ random permutation of $\{ n_p (i-1)+1,\dots , n_p i \}$\;
    $\tilde{\lambda}^{i} \leftarrow$ all permutations of $\{ n_p (i-1)+1,\dots , n_p i \}:\;\; \lbrack \tilde{\lambda}_{1}^{i}, \dots, \tilde{\lambda}_{n_p!}^{i} \rbrack ^\top  $\;
  }\label{endfor}
  $\Lambda \leftarrow$ $\lbrack \lambda_{1}^{\top} \dots \lambda_{n_f}^{\top} \rbrack^\top$ with sub-vectors $\lambda_i^\top$ rearranged in a random order\\
  $F \leftarrow$ all permutations of $\{ 1, \dots, n_f \}:\;\; \lbrack F_{1}, \dots, F_{n_f!} \rbrack^\top$  \\
  \For{$h = 2:N$}
    {
     $D_{max} \leftarrow \emptyset$ \\
     $\Lambda^\prime \leftarrow \emptyset$ \\
     \For{$f = 1:n_f!$}
      {
	\For{$i=1:(n_p!)^{n_f - 1}$}
	{
	  \For{$j=2:n_f$}
	  {
	    $k \leftarrow \left\lceil\frac{\left( \left(i-1\right) \mod \left(n_p! \right)^{j-1} \right)+1}{\left(n_p! \right)^{j-2}}\right\rceil$ \\
	    $\underset{\overline{}}{\tilde{\boldsymbol\lambda}}^{j} \leftarrow \lbrack \tilde{\lambda}_{k}^{j} \dots \tilde{\lambda}_{k}^{j} \rbrack^\top \in \mathbb{R}^{n_p!\times n_p}$
	  }
	  $\Lambda^{\prime\prime} \leftarrow$ arrange $\lbrack\tilde{\lambda}^1 \underset{\overline{}}{\tilde{\boldsymbol\lambda}^2} \dots \underset{\overline{}}{\tilde{\boldsymbol\lambda}^{n_f}} \rbrack^\top$ in order $F_f$ \\
	  $D \leftarrow Hamming(\Lambda,\Lambda^{\prime\prime})$ \\
	  $\overline{D} \leftarrow \frac{1}{h-1}\mathbf{1}^\top D$ \\
	  $D_{max} \leftarrow \begin{bmatrix} D_{max} && \max_{k}\overline{D}_k \end{bmatrix}$ \\
	  $j \leftarrow \operatorname*{argmax}_k \overline{D}_k$ \\
	  $\Lambda^{\prime} \leftarrow \begin{bmatrix} \Lambda^{\prime} && \Lambda^{\prime\prime}_j \end{bmatrix}$ \\
	}
      }
      $j \leftarrow \operatorname*{argmax}_k D_{max(k)}$ \\
      $\Lambda \leftarrow \begin{bmatrix} \Lambda && \Lambda_{j}^{\prime}\end{bmatrix}$
    }
  
  \Return{$\Lambda$}
}
\caption{Sampling Permutations with Spatial Coherence}\label{algo1}
\end{algorithm}

\subsection{Generating Video Jigsaw Puzzles}
We describe here the strategy to generate puzzles from the video frame patches. Noroozi and Favaro \cite{noroozi2016unsupervised} proposed to generate permutations of $9$ image patches by maximizing the Hamming distance between the sampled permutations and the subsequently sampled permutations. They iterate over all possible permutations of $9$ patches till they end up with $N$ permutations; in their case, $N = 100$. In our case, since each video frame is divided into $4$ patches and there are $3$ frames in a tuple, it is not possible to sample permutations from all possible permutations (which is $12!$) due to memory constraints. To reimplement \cite{noroozi2016unsupervised}'s approach, we devise a computationally heavy but memory-efficient means to generate $100$ permutations from $12!$ possibilities. More details on how we generate these permutations are described in the supplementary material. This way, we generate the Hamming-distance based permutations as suggested by \cite{noroozi2016unsupervised}. 

The permutation sampling approach described above treats all video frame patches as one giant image --- thus, the patch belonging to the first frame may get shuffled to the last frame's position (to maximize Hamming distance between the permutations). We treat this permutation sampling approach as an (expensive) baseline but propose another sampling strategy to minimize compute and memory constraints. Our proposed approach can scale to any number of permutations. We generate permutations with a $2\times2$ grid per frame. Our proposed approach forces the sampled permutations not only to obey the Hamming distance criteria but also to respect spatial coherence in video frames. This scales down computational and memory requirements dramatically while giving similar or better performance on transfer learning tasks. Our proposed permutation sampling approach is given in Algorithm~\ref{algo1} and visually presented in Figure 3. 
\vspace{-3mm}
\paragraph{\textbf{Explanation of Algorithm~\ref{algo1}:}} With the constraint of spatial coherence \textit{i.e.} patches within a frame constrained to stay together, the full space of hashes consists of $(n_p!)^{n_f} \times n_f!$ possibilities. After generating the first hash randomly (lines 2, 3 and 5), each next hash $\Lambda_h\;h\in2,\dots N$ is picked by maximizing over the full space the average Hamming distance from previously generated hashes. We divide the full space into subsets of $n_p!$ hashes. Iterating through each subset $\Lambda^{\prime\prime}$ (lines 10-11), we store the best hash from the subset into $\Lambda^\prime$ along with its distance metric into $D_{max}$ (lines 16-20). When the full space is traversed, the best from the good ones ($\Lambda^\prime$) is chosen as the new hash (lines 21-22). Lines 4, 6 and 10-15 describe how each subset $\Lambda^{\prime\prime}$ is constructed. $\Lambda^{\prime\prime}$ contains all $n_p!$ permutations of patches within the first frame but only a particular permutation of patches from the other frames. For memory efficiency, it is sufficient to only create one matrix $\tilde\lambda^1$ that has all patch permutations within the first frame \textit{i.e.} it is not necessary to create $\tilde\lambda^i\;i\in2,\dots,n_f$ as done in line 4. This is because the former is reused in every iteration but only one row from the latter is used to create $\tilde{\boldsymbol\lambda}^i$, the matrix of repeated rows (line 14) which can be achieved by picking the corresponding row from $\tilde\lambda^1$ and adding the offset $n_p(i-1)$ to each element of the row.

\section{Experiments}
In this section we describe in detail our experiments on video action recognition using the video jigsaw network and a comprehensive ablation study, justifying our design choices and conclusions. The datasets we use for training the video jigsaw network are UCF101 \cite{soomro2012ucf101} and Kinetics \cite{kay2017kinetics}. The datasets we evaluate on are UCF101 \cite{soomro2012ucf101} and HMDB51 \cite{kuehne2011hmdb} for video action recognition.

\subsection{Datasets}
UCF101 \cite{soomro2012ucf101} is a benchmark video action recognition dataset consisting of 101 action categories and 13,320 videos; around 9.5k videos are used for training and 3.5k videos are for testing. HMDB51 \cite{kuehne2011hmdb} consists of around 7000 vidoes of 51 action categories, out of which 70\% belong to training set and 30\% are in the test set. Kinetics dataset \cite{kay2017kinetics} is a large scale human action video dataset consisting of 400 action categories and more than 400 videos per action category. 

\subsection{Video Jigsaw Network Training}
\paragraph{\textbf{Tuple Sampling Strategy}}
For our unsupervised pretraining step on UCF101, we use the frame tuples (4 frames/tuple) provided by the authors of \cite{lee2017unsupervised}. They extracted optical flow based regions from these frame tuples and used them in the temporal sequence sorting task \cite{lee2017unsupervised}. We do not use the optical flow based regions from the frames but only use the tuples as a whole. For a given frame tuple $f_1, f_2, f_3, f_4$, we further sample three frames in the following way: \\ $\lbrack (f_1,f_2,f_3), (f_2,f_3,f_4), (f_1,f_3,f_4), (f_1,f_2,f_4) \rbrack$. Hence, we end up with around 900,000 frame tuples from UCF101 dataset to train our video jigsaw network on. In Kinetics dataset, each video is $10$ seconds long. We create our tuples by sampling the $1^{st}$, $5^{th}$ and $10^{th}$ frames from each video. The reason we do not sample further (as we did in the case of UCF101 dataset) is simply that Kinetics dataset is very large and diverse with more than 400 videos per class. This is not true for UCF101 dataset. Note that we do not use any further preprocessing to generate the frame tuples for our video jigsaw network. Previous approaches have used expensive detection and tracking methods \cite{wang2015unsupervised} or optical flow computation to sample the high motion patches \cite{lee2017unsupervised}. 
\vspace{-3mm}
\paragraph{\textbf{Implementation Details}}
We use Caffe \cite{jia2014caffe} deep learning framework for all our experiments and \textit{CaffeNet} \cite{krizhevsky2012imagenet} as our base network, only with $12$ streams for $12$ patches per tuple. Our video jigsaw puzzles are generated on the fly according to the permutation matrix $\Lambda$ generated before training begins. Each row of $\Lambda$ corresponds to a unique permutation of $12$ patches. The video frame patches are shuffled according to the sampled permutation from $\Lambda$ and input to the network. The network is trained to predict the index in $\Lambda$ from which the permutation was sampled. Each video frame is cropped to $224\times224$, then divided into a $2\times2$ grid. Each grid is $112\times112$ pixels and we randomly sample a $64\times64$ patch from it. This strategy ensures that the network can not learn the location of the patches from low level appearance and texture details. We normalize each patch independently from others, to have zero mean and unit standard deviation. This is also done to prevent the network from learning low-level details (also called `network shortcuts' in the self-supervision literature). Each patch is input to the multi-stream video jigsaw network as depicted in Figure~\ref{fig2}. We use a batch size of $150$ and train the network with Stochastic Gradient Descent (SGD) using an initial learning rate of $0.000128$, which decreases by 10 every 128,000 iterations. Each layer in our network is initialized with xavier initialization \cite{glorot2010understanding}. We train the network for 500,000 iterations (approximately 80 epochs) using a Titan X GPU. Our training converges in around 62 hours.  
\vspace{-3mm}
\paragraph{\textbf{Progressive Training Approach}}
We borrow principles from curriculum  learning \cite{bengio2009curriculum} to train our video jigsaw network with an easy jigsaw puzzle task first and then train it for a harder task. We define an easy jigsaw puzzle task as one which has lower $N$ as compared to a harder task as the network has to learn fewer configurations of the patches in the video frames. So instead of starting from scratch for say, $N = 500$, we initialize the network's weights with the weights of the network with $N = 250$. 

\vspace{-3mm}
\paragraph{\textbf{Avoiding Network Shortcuts}}
As mentioned in recent self-supervised approaches \cite{doersch2015unsupervised,noroozi2016unsupervised,noroozi2017representation}, it is imperative to deal with the self-supervised network's tendency to learn the patch locations via low level details such as due to chromatic aberration. Typical solutions to this problem are channel swapping \cite{lee2017unsupervised}, color normalization \cite{noroozi2016unsupervised}, leaving a gap between sampled patches and training with a percentage of images in grayscale rather than color \cite{noroozi2017representation}. All these approaches aim to make the patch location learning task harder for the network. Our video jigsaw network incorporates these techniques to avoid network shortcuts. Our patch size is kept $64 \times 64$ sampled from within a $112 \times 112$ window. Around half of the total video frames are randomly projected to grayscale and we normalize each sampled patch independently. Our experiments using these techniques result in a drop in performance in video jigsaw puzzle solving accuracy but the transfer learning accuracy increases.

\vspace{-3mm}
\paragraph{\textbf{Choice of Video Jigsaw Training Dataset}}
As mentioned, we train video jigsaw networks using UCF101 and Kinetics datasets. Our results using the two datasets are shown in Table~\ref{T1}. We show video jigsaw task accuracy (VJ Acc) and the finetuning accuracy on UCF101 (Finetune Acc) for pretraining with both datasets. $N$ is the number of permutations. We can note two things from the table. Using Kinetics results in a worse video jigsaw solving performance, but results in a better generalization and transfer learning. Our finetuning results are consistently better with Kinetics pretraining as compared to training on UCF101. This shows that a large-scale diverse dataset like Kinetics is able to generalize to a completely different dataset (UCF101). One possible reason behind the reduced performance of UCF101 dataset is the fact that we oversample from it. This results in an easy task for the video jigsaw network to learn the low-level details of the video frame appearances and rapidly decrease the training loss. However, this would not result in a good transfer learning performance. To test this hypothesis, we use the reduced version of the UCF101 dataset (without any oversampling), comprising just 200,000 frame tuples and train video jigsaw networks for $N = 500$ and $N = 1000$. The results are shown in Table~\ref{T1-2}. As is shown, even without oversampling, UCF101-based pretraining does not perform as well as Kinetics dataset.

\begin{table}[t]
\centering
\makegapedcells

\resizebox{0.5\textwidth}{!}{%
\begin{tabular}{|l|r|r|r|r|}
\hline
\textbf{Pretraining Dataset} & \textbf{\begin{tabular}[c]{@{}l@{}}VJ Acc (\%)\\       (N = 100)\end{tabular}} & \textbf{\begin{tabular}[c]{@{}l@{}}Finetune Acc (\%)\\(N = 100)\end{tabular}} & \textbf{\begin{tabular}[c]{@{}l@{}}VJ Acc (\%)\\ (N = 250)\end{tabular}} & \textbf{\begin{tabular}[c]{@{}l@{}}Finetune Acc (\%)\\(N = 250)\end{tabular}} \\ \hline
\textbf{UCF101} & 97.6 & 44.0 & 84.6 & 42.6 \\  \hline
\textbf{Kinetics} & 61.6 & \textbf{44.6} & 44.0 & \textbf{49.0} \\ \hline
\end{tabular}%
}
\caption{Comparison between UCF101 and Kinetics datasets for video jigsaw training}
\label{T1}

\end{table}

\begin{table}[h]
\makegapedcells
\centering
\resizebox{0.5\textwidth}{!}{%
\begin{tabular}{|l|r|r|r|r|}
\hline
\textbf{Pretrained On} & \textbf{VJ Acc (\%)} & \textbf{Finetune Ac (\%)} & \textbf{VJ Acc (\%)} & \textbf{Finetune Ac (\%)} \\ \hline
\textbf{Kinetics} & 40.3 & \textbf{49.2} & 29.4 & \textbf{54.7} \\ \hline
\textbf{UCF101-no oversampling} & 63.3 & 46.5 & 58 & 46.4 \\ \hline
\textbf{N = no. of permutations} & N = 500 & N = 500 & N = 1000 & N = 1000 \\ \hline
\end{tabular}%
}
\caption{Comparison between Kinetics and the original UCF101 frame tuples as pretraining dataset for video jigsaw network}

\label{T1-2}
\end{table}

\vspace{-3mm}
\paragraph{\textbf{Choice of Number of Permutations}}
We vary the number of permutations $N$ a video jigsaw network has to learn. We start with $N = 100$ and take it up to $N = 1000$. As we increase the number of permutations (see Table~\ref{T2}), the network finds it harder to learn the configuration of the patches, but the generalization improves. This experiment is run with Kinetics dataset trained on video jigsaw network. 

\begin{figure}[b!]
\centering
\includegraphics[width=1.0\linewidth,height=6.6cm,keepaspectratio]{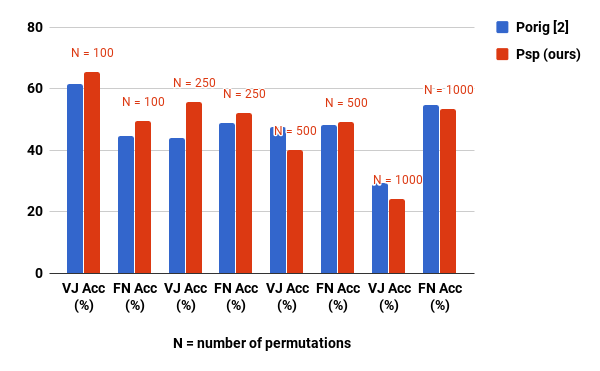}
\caption{Comparison between the permutation strategy proposed by \cite{noroozi2016unsupervised} ($P_{orig}$) and our proposed sampling approach ($P_{sp}$) on the video jigsaw task (indicated by VJ Acc) and the finetuning task on UCF101 (indicated by FN Acc) for various different number of permutations $N$. Our approach consistently performs better or comparable to the approach of \cite{noroozi2016unsupervised} while saving memory and computational costs. Figure is best viewed in color}
\label{fig4}
\end{figure}

\begin{table}[t]
\makegapedcells
\centering
\resizebox{0.4\textwidth}{!}{%
\begin{tabular}{|r|r|r|}
\hline
\textbf{No. of permutations} & \textbf{VJ Acc (\%)} & \textbf{Finetuning Ac (\%)} \\ \hline
100 & 61.6 & 44.6 \\ \hline
250 & 44.0 & 49.0 \\ \hline
500 & 47.6 & 48.1 \\ \hline
1000 & 29.4 & \textbf{54.7} \\ \hline
\end{tabular}%
}
\caption{As we increase $N$, the video jigsaw performance decreases but the finetuning accuracy increases}
\label{T2}

\end{table}
\vspace{-3mm}
\paragraph{\textbf{Permutation Generation Strategy}}
We compare the performance of our proposed permutation strategy which enforces spatial coherence (referred to as $P_{sp}$) between permuted patches --- with the proposed approach of \cite{noroozi2016unsupervised} (referred to as $P_{orig}$). We show results for this comparison in Figure~\ref{fig4}. As the bar chart shows, for various number of permutations, our proposed spatial coherency preserving method either outperforms the original random permutation generation strategy or is comparable to it, while being many times faster to generate. 

\vspace{-3mm}
\paragraph{\textbf{Patch Size}}
We also compare the performance of our video jigsaw method trained on different frame patch sizes. Table ~\ref{T2aa} shows that the finetuning accuracy increases with the increase in patch size but does not give much improvement beyond a patch size of $80 \times 80$. 

\begin{table}[t]
\centering
\resizebox{0.3\textwidth}{!}{%
\begin{tabular}{|r|r|r|}
\hline
\textbf{Patch size} & \textbf{Finetuning Ac (\%)} \\ \hline
64 & 54.7 \\ \hline
80 & \textbf{55.4} \\ \hline
100 & 54.1 \\ \hline
\end{tabular}%
}
\caption{As we increase patch size, the video jigsaw finetuning accuracy on UCF101 dataset increases}
\label{T2aa}
\end{table}

\subsection{\textbf{Finetuning for Action Recognition}}
Once the video jigsaw network is trained, we use the convolutional layers' weights to initialize a standard \textit{CaffeNet} \cite{krizhevsky2012imagenet} architecture and use it to finetune on UCF101 and HMDB51 datasets. For UCF101, we sample 25 equidistant frames per video and compute frame-based accuracy as our finetuning evaluation measure. For HMDB51 we sample 1 frame per second from each video and use them for the finetuning experiment. With our best model and parameters (pretrained on Kinetics dataset), results are given in Table~\ref{T3} for test split 1 of both UCF101 and HMDB51 datasets.  

\begin{table}[t!]
\makegapedcells

\centering
\resizebox{0.5\textwidth}{!}{%
\begin{tabular}{@{}lrr@{}}
\toprule
\textbf{Pretraining} & \textbf{UCF101 Acc (\%)} & \textbf{HMDB51 Acc (\%)} \\ \midrule
random & 40.0 & 16.3 \\ \midrule
ImageNet (with labels) & 67.7 & 28.0 \\
Fernando \etal \cite{fernando2017self} & 60.3 & 32.5 \\ \midrule
Hadsell \etal \cite{hadsell2006dimensionality} & 45.7 & 16.3 \\
Mobahi \etal \cite{mobahi2009deep} & 45.4 & 15.9 \\
Wang and Gupta \cite{wang2015unsupervised} & 40.7 & 15.6 \\
Misra \etal \cite{misra2016shuffle} & 50.9 & 19.8 \\
Lee \etal \cite{lee2017unsupervised} & 56.3 & 22.1 \\
Vondrick \etal \cite{vondrick2016generating} & 52.1 & - \\
Video Jigsaw Network (ours) & \textbf{55.4} & \textbf{27.0} \\ \bottomrule
\end{tabular}%
}
\caption{Finetuning results on UCF101 and HMDB51 of our proposed video jigsaw network (pretrained on Kinetics dataset with $N = 1000$ permutations --- compared to the state of the art approaches. Note that all these results are computed using \textit{CaffeNet} architecture. Our method gives superior or comparable performance to the state of the art unsupervised learning + finetuning approaches that use RGB frames for training}
\label{T3}
\end{table}

Table~\ref{T3} shows our video jigsaw pretraining approach outperforming recent unsupervised pretraining approaches when finetuning on HMDB51 dataset. On UCF101 dataset, our finetuning accuracy is comparable to the state of the art. The method of Fernando \etal uses a different input from ours (stacks of frame differences) whereas we use RGB frames to form the jigsaw puzzles. All other approaches operate on RGB video frames or frame patches hence we can fairly compare with them. The methods of Lee \etal \cite{lee2017unsupervised} and Misra \etal \cite{misra2016shuffle} are pretrained on UCF101 dataset whereas our best network is trained on Kinetics dataset. This again shows the domain transfer capability of a large scale dataset like Kinetics, compared to UCF101. Our method achieves this without doing any expensive tracking \cite{wang2015unsupervised} or optical flow based patch or frame mining such as \cite{misra2016shuffle,lee2017unsupervised}. This means that our approach requires large scale diverse unlabeled video dataset to work. We used $3$ frames per video from Kinetics dataset --- hence we were only using about 400,000 tuples for our video jigsaw training. We believe that using a larger dataset would lead to better performance, given that our approach is close to the state of the art. Another point to note is that methods which perform well on UCF101 such as Lee \etal \cite{lee2017unsupervised} and Misra \etal \cite{misra2016shuffle} do not perform that well on HMDB51, whereas our method actually generalizes well, given that it is pretrained on a completely different dataset. 

\begin{table}[ht!]
\centering
\resizebox{0.5\textwidth}{!}{%
\begin{tabular}{@{}llr@{}}
\toprule
\textbf{Method} & \multicolumn{1}{l}{\textbf{Supervision}} & \multicolumn{1}{l}{\textbf{Classification}} \\ \midrule
ImageNet & 1000 class labels & 78.2\% \\ \midrule
Random \cite{pathak2016context} & none & 53.3\% \\
Doersch \etal \cite{doersch2015unsupervised} & ImageNet context & 55.3\% \\
Jigsaw Puzzle \cite{noroozi2016unsupervised}& ImageNet context & 67.6\% \\
Counting \cite{noroozi2017representation} & ImageNet context & 67.7\% \\ \midrule
Wang and Gupta \cite{wang2015unsupervised} & 100k videos, VOC2012 & 62.8\% \\
Agrawal \etal \cite{agrawal2015learning} & egomotion (KITTI, SF) & 54.2\% \\
Misra \etal \cite{misra2016shuffle} & UCF101 videos & 54.3\% \\
Lee \etal \cite{lee2017unsupervised} & UCF101 videos & 63.8\% \\
Pathak \etal \cite{pathak2017learning} & MS COCO + segments & 61.0\% \\
Video Jigsaw Network (ours) & Kinetics videos & \textbf{63.6\%} \\ \bottomrule
\end{tabular}%
}
\caption{PASCAL VOC 2007 classification results compared with other methods. Other results taken from \cite{noroozi2017representation} and \cite{lee2017unsupervised}}
\label{T2221}

\end{table}

\subsection{Results on PASCAL VOC 2007 Dataset}
The PASCAL VOC 2007 dataset consists of 20 object classes with 5011 images in the train set and 4952 images in the test set. Multiple objects can be present in a single image and the classification task is to detect whether an object is present in a given image or not. We evaluate our video jigsaw network on this dataset by initializing a \textit{CaffeNet} with our video jigsaw network's trained convolutional layers' weights. The fully-connected layers' weights are randomly sampled from a Gaussian distribution with zero mean and 0.001 standard deviation. Our finetuning scheme follows the one suggested by \cite{krahenbuhl2015data}. Our classification results on the Pascal VOC 2007 test set are shown in Table~\ref{T2221}.

Our trained network generalizes well not only across datasets but also across tasks. Our video jigsaw network is trained on Kinetics videos and not on object-centric images, yet performs competitively against the state-of-the-art image-based semi-supervised approaches and outperforms most of the video-based semi-supervised methods. 

\subsection{Visualization Experiments}
\begin{figure}[t]
\centering
\includegraphics[height=2.7cm]{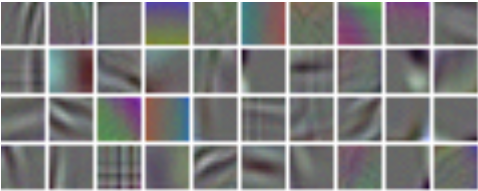}
\caption{Visualization of first 40 learned conv1 filters of our best performing video jigsaw model}
\label{fig111}
\end{figure}

\begin{figure}[h!]
\centering
\includegraphics[height=5.1cm,keepaspectratio]{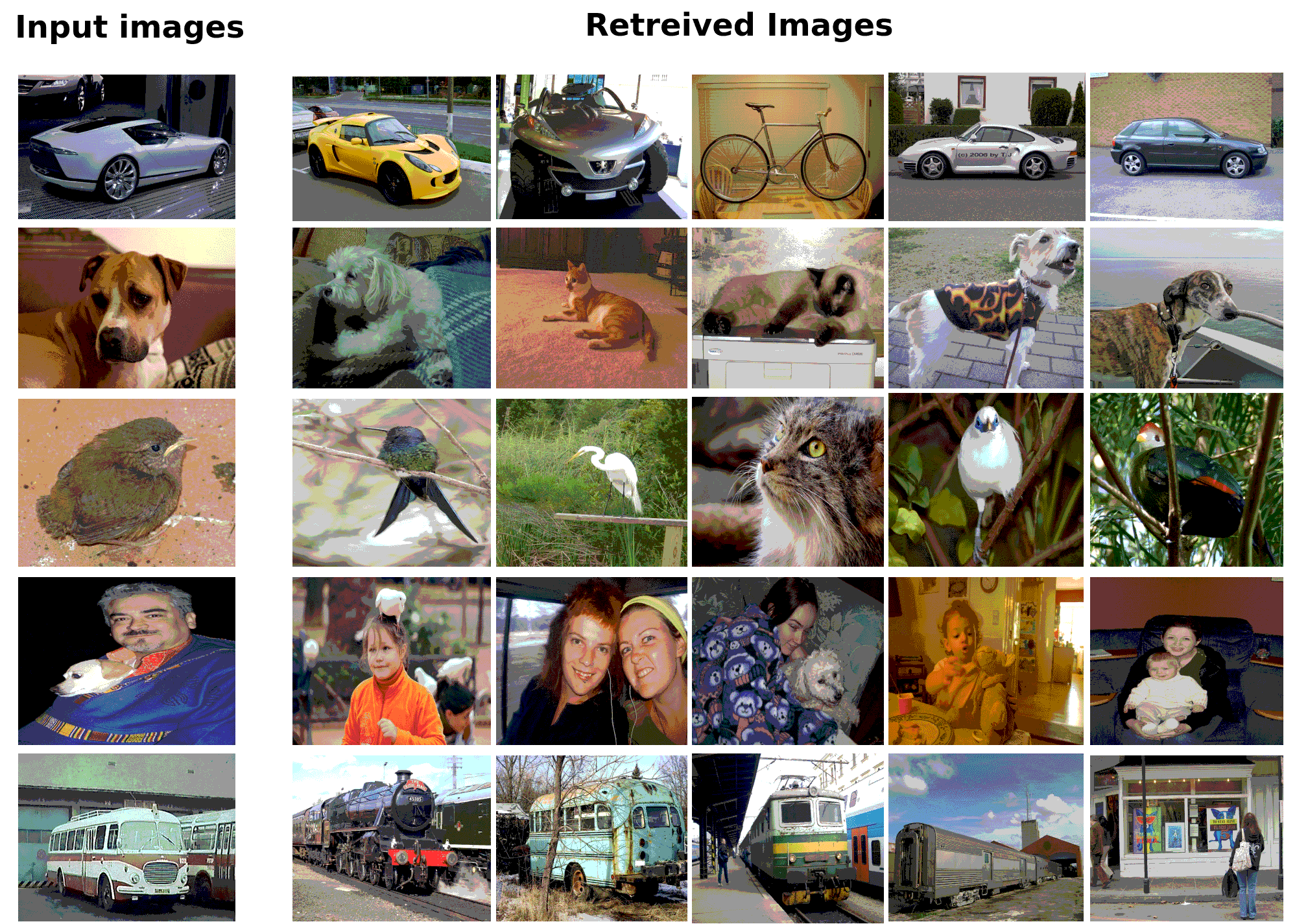}
\caption{Retrieval Experiment on PASCAL VOC dataset using our model}
\label{fig112}
\end{figure}

We show first 40 conv1 filter weights of our best video jigsaw model in Figure~\ref{fig111} which show oriented edges learned by our model. Note that training this model does not use activity labels. We also perform a qualitative retrieval experiment on the video jigsaw model finetuned on Pascal VOC dataset. Results are shown in Figure~\ref{fig112}. We note that the retrieved images returned by the model match the query image which qualitatively shows that our model trained on unlabeled videos is able to identify objects in still images. 

\section{Conclusion}
We propose a self-supervised learning task where spatial and temporal contexts are exploited jointly. Our framework is not dependent on heavy preprocessing steps such as object tracking or optical flow based patch mining. We demonstrate via extensive experimental evaluations that our approach performs competitively on video activity recognition, outperforming the state of the art in self-supervised video action recognition on HMDB51 dataset. We also propose a permutation generation strategy which respects spatial coherency and demonstrate that even for shuffling $12$ patches, diverse permutations can be generated extremely efficiently via our proposed approach. 

{\small
\bibliographystyle{ieee}
\bibliography{egbib}
}

\end{document}